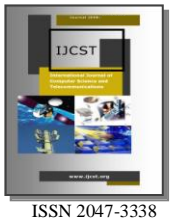

# The State of the Art Recognize in Arabic Script through Combination of Online and Offline

Dr. Firoj Parwej

Department of Computer Science, Jazan University, Jazan, Kingdom of Saudi Arabia (KSA)

dr.firojparwej@gmail.com

*Abstract*— Handwriting recognition refers to the identification of written characters. Handwriting recognition has become an acute research area in recent years for the ease of access of computer science. In this paper primarily discussed On-line and Off-line handwriting recognition methods for Arabic words which are often used among then across the Middle East and North Africa People. Arabic word online handwriting recognition is a very challenging task due to its cursive nature. Because of the characteristic of the whole body of the Arabic script, namely connectivity between the characters, thereby the segmentation of An Arabic script is very difficult. In this paper we introduced an Arabic script multiple classifier system for recognizing notes written on a Starboard. This Arabic script multiple classifier system combines one off-line and on-line handwriting recognition systems. The Arabic script recognizers are all based on Hidden Markov Models but vary in the way of preprocessing and normalization. To combine the Arabic script output sequences of the recognizers, we incrementally align the word sequences using a norm string matching algorithm. The Arabic script combination we could increase the system performance over the excellent character recognizer by about 3%. The proposed technique is also the necessary step towards character recognition, person identification, personality determination where input data is processed from all perspectives.

## I. INTRODUCTION

THE handwritten automatic recognition has been classified into two types of categories based on the input data offline and online. The off-line handwriting recognition involves the automatic conversion of text in an image into letter codes which are usable within computer and text-processing applications. The data obtained from this form is regarded as a static representation of handwriting [1]. The off-line handwriting recognition is comparatively difficult, as different people have different handwriting styles. The root of online handwriting recognition is real time data collection by way of a digital sampling method [2].

The on-line handwriting recognition involves the automatic conversion of text as it is written on a special digitizer, PDA or digitizing tablets, where a sensor picks up the pen-tip movements as well as pen-up/pen-down switching. That kind of data is known as digital ink and can be regarded as a dynamic representation of handwriting. The obtained signal is converted into letter codes which are usable within computer and text-processing applications. The on-line method has [3] the following advantages compared to the off-line method firstly the memory requirements for the dictionary are significantly smaller than offline and secondly the recognition speed is faster. In this paper we consider a novel task, which is the recognition of text written on a Starboard. We are using the multiple cameras, microphones, Hitachi Starboard, electronic pens for note-taking, a projector, and an electronic Starboard. We consider an on-line recognition problem, namely the recognition of notes written on a Starboard.

The field of character recognition for Arabic scripts faces major problems. These problems are due to the complexities of this script like cursiveness, multiple shapes [4] of one character at different positions in a ligature, overlapping and connectivity of characters on the baseline. In the present paper we describe the combination of three different systems derived from the recognizers [5]. To combine the output of the recognizers, we incrementally align the word sequences using a standard string matching algorithm. After that the word that most often occur at a certain position is used as the final result. As Arabic script base languages are very complex as compared to the other languages thus it is very difficult to obtain a suitable precision only faithfully on the online data. Finally the results in an overall performance of about 68.2%.

## II. ARABIC SCRIPT SYSTEM FRAMEWORK

The Hitachi Starboard interface is used for recording the Arabic script handwriting. This classroom is able to record Arabic scripts with up to one participant. It is equipped with multiple cameras, microphones, electronic pens for note-taking, a projector, and an electronic whiteboard. A schematic overview of this classroom is presented in Fig. 1. The camera is located in the middle of the Starboard schema.

The board is typically mounted to a wall or floor stand. An interactive Starboard (ISWB) device is connected to a computer via USB or a serial port cable, or else wirelessly via Bluetooth or a 2.4 GHz wireless. In the latter case WEP and WPA/PSK security is available. Movement of the user's finger, pen, or other pointer over the image projected on the whiteboard is captured by its interference with infrared light at the surface of the whiteboard.

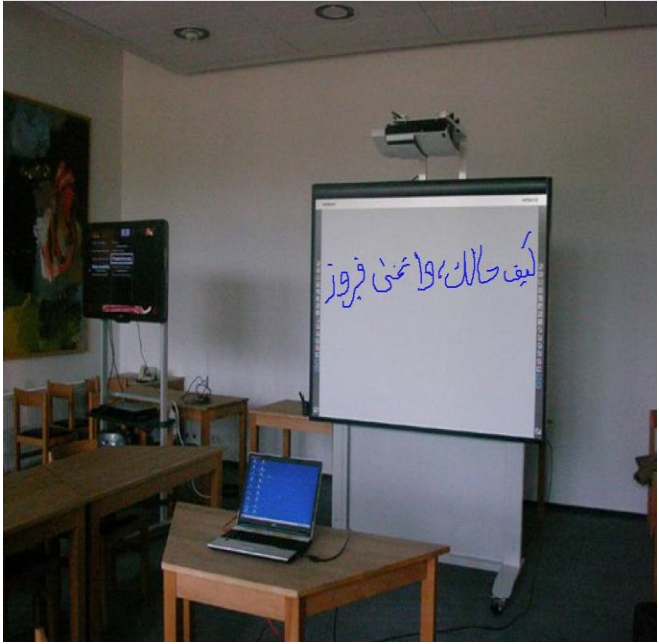

Fig. 1: The Hitachi Starboard Interface Recording

The handwritten data may be present in online or offline format. In the case of online recognition, a time ordered sequence of coordinates, representing the pen movement, is available [6]. The acquisition interface outputs a sequence of (x, y) coordinates representing the location of the spike of the pen together with a time stamp for each location. This may be produced by any electronic sensing device, such as a mouse, an electronic pen on a tablet, or a camera recording gestures. In the case of offline recognition, only the image of the text is present, which usually is scanned or photographed from paper. The frame rate of the recordings varies from 25 to 80 frames per second.

This portion has shown figure 2 a model for the Arabic script recognition system. In this model first section is preprocessing, where the input is raw Arabic handwritten data and the output usually consists of extracting text lines. If the Arabic data have been acquired from a system that does not produce any noise and only single words have been recorded. If usually the data contains noise which needs to be removed to improve the quality of the handwriting. The preprocessing may also include word extraction and even character segmentation [7].

The task of Arabic scripts character segmentation is especially difficult if considered separately. This is because a word often can not be correctly segmented before it has been recognized, and cannot be recognized without previously segmenting it into Arabic characters. This phenomenon is known as Sayre's paradox [Sayre (1973)]. After preprocessing phase, the recorded data need to be normalized. This is a very important part of Arabic handwriting recognition system, because the writing styles of the writers differ with respect to skew, slant, height, and width of the Arabic characters. The text line is corrected in regard to its skew i.e., It is rotated, so that the baseline is parallel to the x-axis. Then, slant correction is performed so that the slant becomes upleft.

The Feature extraction is an important part of any handwriting recognition system, for any classifier needs numeric data as input. We are using a common method in offline recognition of handwritten text lines is the use of a sliding window moving in the writing direction over the text. After preprocessing, normalization and feature extraction the same methods can be used for offline and online recognition. The next phase is the classification of the input features. This classifier based on Hidden Markov Models is applied [8]. The last recognition phase is the post processing. In this phase only possible if additional knowledge about the domain is available.

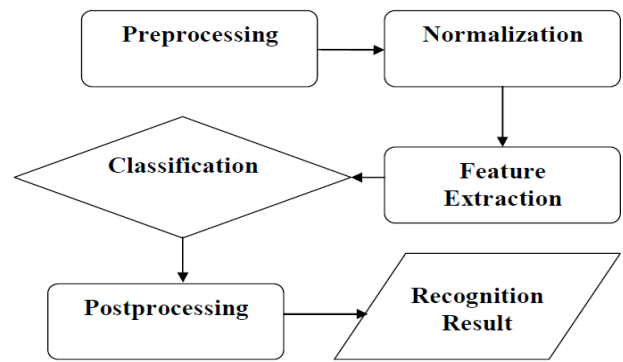

Fig. 2: The Arabic Scripts Recognition System

### III. PROPOSED ONLINE AND OFFLINE SYSTEM FOR PREPROCESSING

The Pre-processing of the raw input strokes is crucial for the success of efficient character recognition systems. The aim of pre-processing step is to process the input strokes so that variation can be erased and forward it to the next process for accurate feature extraction and for achieving better efficient recognition rate for Arabic script. In this paper we are using different preprocessing steps on the input strokes from both online and offline likelihood to better utilize in possible for recognition. In the online different type of preprocessing steps [9] are involved in stroke acquisition, stroke segmentation, smoothing, interpolation and de-hooking. In the different type of offline preprocessing steps are involved in stroke combining, baseline finding.

*A. Online*

The online handwriting recognition means that the machine recognizes the writing while the user writes. The term real time or dynamic has been used in place of online. It's depending on the recognition technique and the speed of the computer, the recognition lags behind the writing to a greater or lesser extent. The input strokes are obtained through the tablet pc or digital pen. These strokes consist of x and y coordinates, coordinate representing the location of the spike of the pen together with a time stamp for each location which will be used for preprocessing. In interpolation facing the limiting processing power of pen and low camera frame rate, it skips some points this depends upon the writing speed, to compute the missing point's interpolation is performed for correct classification. These types of problem solved by using Bresenham's line algorithms.

Sometimes when the user input strokes contain zigzag path due to naturally hand shivering during writing. Such type of the slow variation, but this may effect on the Arabic scripts recognition rate. Smoothing is one of the simplest approaches for data filtering. It consists of substituting the coordinates of the original point by using the neighboring points to solve the problem. Ours pen is very sensitive when you are fast writing or writing by inexperienced people, then pen-up and pen-down events are generated. This type of problem called in the De-hooking, it is very common artifacts found at the beginning and end of the strokes. These often create problems in the recognitions of the Arabic scripts. Therefore, it is very important to remove them. The hooks occurring at the beginning and the end of the stroke are isolated with the help of the generated chain codes [10].

*B. Offline*

In this phase offline handwriting recognition involves the automatic conversion of text in an image into letter codes which are usable within computer and text-processing applications. The data obtained from this form is regarded as a static representation of handwriting. The off-line handwriting recognition is comparatively difficult, as different people have different handwriting styles. The offline character recognition often involves scanning a form or document written sometime in the past. This means the individual characters contained in the scanned image will need to be extracted. After the extraction of individual characters occurs, a recognition engine is used to identify the corresponding computer character.

In only online finding are not enough for Arabic online character recognition due to its characteristics. It is more suitable to involve offline information along with online information to increase the recognition rate. In digraph combination, it is difficult to write some digraphs i.e., ☐ without lifting the pen while online character recognition does not permit to lift a pen during writing. The baseline is the imaginary horizontal line with which the base of each character, excluding descenders, is aligned. It also corresponds to the bottom of the x-height. The baseline information has been used for different purposes in Arabic script handwriting recognition. The baseline represents a first orientation in a word. This line represents an orientation in a word and is necessary for many different handwriting task i.e., Personality identification, writer identification.

## IV. RECOGNITION FOR ARABIC SCRIPT

In this paper we are proposing recognition engine in both online and offline system. A Hidden Markov Models is built for each of the 28 Arabic characters in the character set, which includes each character has 2-4 forms depending on its position within the word as well as many letters of the Arabic alphabet have dots, above or below the character body, and some letters have a Hamza (zigzag shape) and Madda. Some Arabic's characters become over each other horizontally when they connected with each other. We are using Hidden Markov Models as the sequential topology, there are only two transitions per state, one to recursive and one in the next state. In the emitting states, the observation likelihood distributions are estimated by continuous Hidden Markov Models is used. According to [S.Gunter, H.Bunke] If the number of Gaussians and training iterations have an effect on the recognition results of an HMM recognizer. If repeat many times of optimal value increase with the amount of training data because more variations are encountered [11]. In this paper we are trained with up to 48 Gaussian components and the classifier that performed better on a validation set. We are proposing for training phase the Baum-Welch algorithm and for recognition phase the Viterbi algorithm.

*A. The Baum-Welch Algorithm*

The Baum-Welch algorithm is a GEM algorithm. In other words GEM means generalized expectation maximization. It can compute maximum likelihood estimations and posterior mode estimations for the parameters (transition and emission probabilities) of a Hidden Markov Models, when given emission as the training data only. The algorithm has two Processes. In the First process, the calculating the forward probability and the backward probability for each Hidden Markov Models state [12]. In the Second process, calculating the transition-emission pair values and multiplied by the probabilities of the whole observation sequence. In another word,

Given $O = (O_1, O_2,…, O_T)$

Estimate $\lambda = (A, B, \pi)$ to maximize $P(O|\lambda)$

The procedures of Baum-Welch Algorithm will be:

• Let the initial model be $\lambda_0$

• Compute new $\lambda$ based on $\lambda_0$ and observation O

• If $\log P(O|\lambda) - \log P(O|\lambda_0) < \vartheta$, then stop

• Else set $\lambda_0 \leftarrow \lambda$ and go to the second step

In the mathematical procedure how this algorithm works.

Define $\xi(i, j)$ as the probability [13] of being in the state i at time t in state j at time t+1.

$$\xi(i,j) = \frac{\alpha_t(i) a_{ij} b_j(o_{t+1}) \beta_{t+1}(j)}{P(O|\lambda)} = \frac{\alpha_t(i) a_{ij} b_j(o_{t+1}) \beta_{t+1}(j)}{\sum_{i=1}^{N}\sum_{j=1}^{N} \alpha_t(i) a_{ij} b_j(o_{t+1}) \beta_{t+1}(j)}$$

Define $\gamma_t(i)$ as the probability of being in the state i at time t, given the observation sequence.

$$\gamma_t(i) = \sum_{j=1}^{N} \xi_t(i,j)$$

Is the expected number of times state i is visited.

$$\sum_{t=1}^{T} \gamma_t(i)$$

Is the expected number of transitions from state i to state j.

$$\sum_{t=1}^{T-1} \xi_t(i,j)$$

The Update rules of Baum-Welch Algorithm

$\pi_i$ = expected frequency in the state i at a time (t = 1) = $\Upsilon 1(i)$

$a_{ij}$ = (expected number of times in state j and observing symbol k) / (expected number of times in state j).

$$a_{ij} = \frac{\sum_{t=1}^{T-1} \xi_t(i,j)}{\sum_{t=1}^{T-1} \gamma_t(i)}, 1 \le i \le N, 1 \le j \le N$$

$b_j(k)$ = (expected number of times in state j and observing symbol k) / ( expected number of times in state j ).

$$b_j(k) = \frac{\sum_{t, o_t = k} \gamma_t(j)}{\sum \gamma_t(j)}$$

In the training process, each Hidden Markov Model is trained for different subject. Each subject needs to have his/her own trained file in order to mend the accuracy for is hidden states decoding [14]. The number of states N is set to be 8. For each training process, the value of P is not fixed, as each training process may take different time, the value of P may transmogrify.

*B. The Viterbi Algorithm*

The Viterbi Algorithm (VA) was first proposed as a solution to the decoding of convolutional codes by Andrew J. Viterbi in 1967, with the idea being further developed by the same author in [15]. The Viterbi algorithm is used to find the most probable activity series (hidden states of HMM) from observation series. The Viterbi algorithm is a dynamic programming algorithm to find the Viterbi path that is hidden in the observed sequence. The Viterbi path can also be regarded as the most likely sequence of hidden states. Assume δ to be the probability of the most probable path to the state. Then $\delta_t(i)$ is the maximum probability of all possible sequences ending at state i when it is time t. The Viterbi path is the sequence which results in this maximal probability [16]. A is the matrix of state-transition probabilities with elements $a_{ij}$, B is the matrix of observation probabilities with elements $b_{ij}$, and π is the vector of initial state probabilities with elements π(i). In mathematical expression how this algorithm works.

Find the path ($q_1$,…, $q_t$ ) that maximizes the likelihood P($q_1$,…, $q_t$ |O, λ)

The Solution by dynamic Programming define here

$$\delta_t(i) = \max_{q_1, q_2, \ldots, q_{t+1}} P(q_1, q_2 \ldots, q_t = i, o_1, o_2, \ldots, o_t | \lambda)$$

By induction we have

$$\delta_{t+1}(j) = \max_i [\delta_t(i) a_{ij}] b_j(o_{t+1})$$

By Initialization we have

$$\delta_i(i) = \pi_i b_i(o_1), 1 \le i \le N$$

$$\varphi_1(i) = 0$$

By Recursion we have

$$\delta_t(j) = \max_{1 \le i \le N} [\delta_{t-1}(i) a_{ij}] b_j(o_t)$$

$$\varphi_t(j) = \max_{1 \le i \le N} [\delta_{t-1}(i) a_{ij}], 2 \le t \le T, 1 \le j \le N$$

By Termination we have:

$$P^* = \max_{1 \le i \le N} [\delta_T(i)]$$

$$q_T^* = \arg\max_{1 \le i \le N} [\delta_T(i)]$$

By Path (state sequence) backtracking we have:

$$q_t^* = \varphi_{t+1}(q_{t+1}^*), t = T-1, T-2, \ldots, 1$$

Finally, the end Viterbi path will be calculated.

## V. ARABIC SCRIPT COMBINATION

In Arabic script improved recognition performance by combining on-line and off-line classifiers. By combining the on-line method with the off-line method, the recognition accuracy is improved because they compensate their disadvantages reciprocally. There exist various possibilities to combine outputs from multiple classifiers. Kittler et al. present many combination schemes, such product rule, majority voting, sum rule, min rule, max rule and median rule [17].

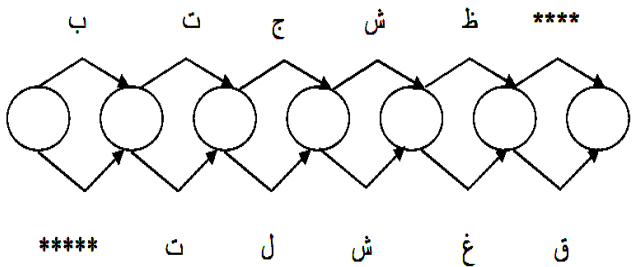

Fig. 3: The ROVER Construction of Arabic Word Transition Network

The recognizers output Arabic script series of whole Arabic script lines and there may be a different number of words in each output series, these series are aligned into an Arabic Script Transition Network (ASTN).

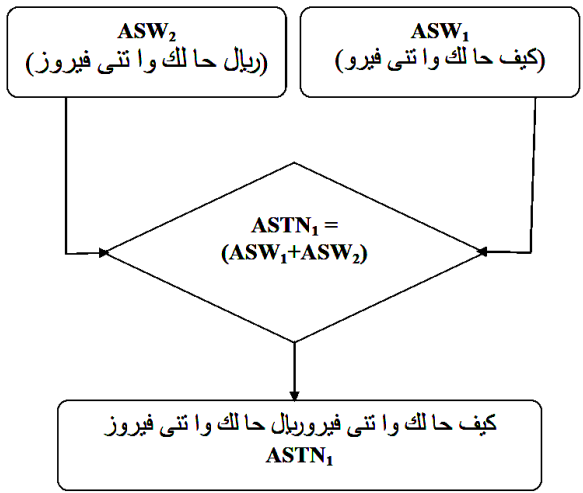

Fig. 4: The Arabic Script Transition Network Aligning Multiple Results Sequence First

The recognizer hypotheses are more robust at higher levels of the system because a wider temporal context is taken into account. The combination of the word level might be more effective than state-level combination. In order to combine word recognition hypotheses we use the Recognizer Output Voting Error Reduction (ROVER) algorithm. The ROVER merges the best Arabic script series output by different recognizers into a new word transition network by dynamic programming alignment [18]. The Arabic words that have been aligned form parallel paths in an Arabic word transition network shown in Fig. 3. The resulting network is then evaluated by a voting module, which, at each node in the network [19], selects the best path based on simple majority voting.

When applying this method to our recognition, we found that some problems were created by the dynamic programming procedure [20]. But we resolve the word transition network by aligning word hypotheses according to their actual time stamps in order to ensure that parallel paths in the transition network represent genuine alternatives covering the same temporal portion of the signal [21]. These suboptimal solutions often provide good alignment precision in rehearse. After the alignment a voting module extracts the best scoring Arabic Script sequence from the ASTN. Another condition if the situation is equals the output of the best performing system on the validation set is taken. An example of an alignment of the output of our four recognizers [22] (denoted by $ASW_1$, $ASW_2$, $ASW_3$ and $ASW_4$) and the final output كيف حا لك وا تنى فيرو in figure four to six sequences.

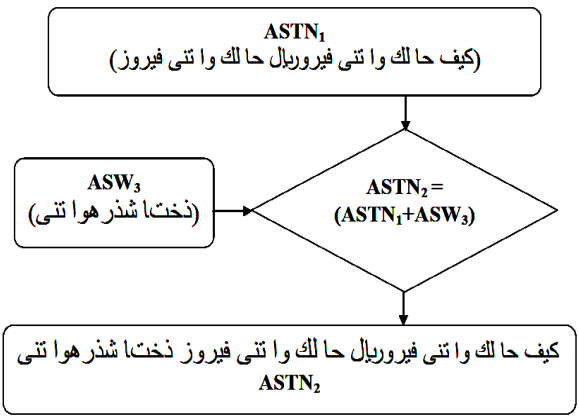

Fig. 5: The Arabic Script Transition Network Aligning Multiple Results Sequence Second

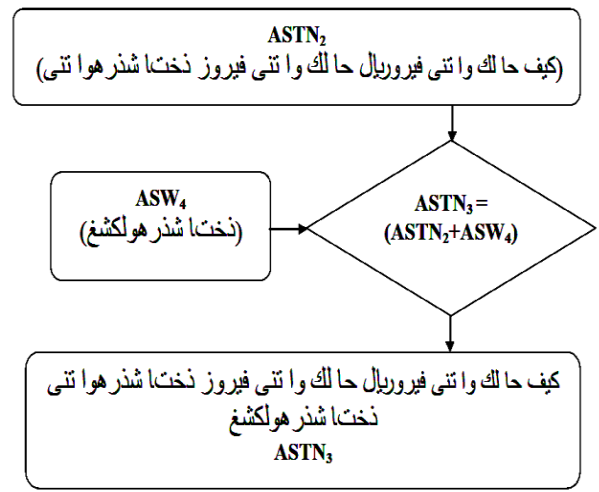

Fig. 6: The Arabic Script Transition Network Aligning Multiple Results Sequence Third

## VI.   EXPERIMENTAL RESULTS

We made experiments to evaluate our proposed combined recognizer for Arabic script recognition. In Arabic scripts first impact of how difficult the reading of whiteboard Arabic scripts and the quality of handwriting on a whiteboard is lower than on-line handwriting presented on an electronic writing tablet or offline handwriting scanned in from paper. Few interruptions occur if we writing with a normal pen on paper or even with an electronic pen on writing tablet than to writing on a whiteboard. Another interruption when we are using a normal pen on paper or an electronic pen on a tablet, the Arabic scripts writer's arm usually rests on a table.

But we are writing on a whiteboard the arm does not rest on any surface any surface, which puts much more stress on the Arabic scripts writers' hand. After that we must expect more noise and deformations in whiteboard handwriting than in normal on-line or off-line handwritten an Arabic script data. The exhaustive amount of the recorded data is 5056 words in 1023 lines from 5 different participants. The recognizer has been dynamically trained with 48 iterations.

In our experimental data is split into four stages, first stage for training second stage for warrantable the meta parameters of the training a second warrantable stage which can be used, the third stage for renovate a language model and last an independent test stage. The table 1 shows the results of the two individual recognition systems in offline and online Arabic scripts. The word recognition rate is simply measured by dividing the number of immaculate recognized Arabic words by the number of words in the transcription.

In Table 2 Experimental results show that a combination of online and offline achieve recognition rates about 68.2%, which is significantly higher than the about 66.3% using a previously developed individual offline and online Arabic script recognition system and the recognition precision could be increased by 1.9 %.

Table 1: The Two Individual Recognition Offline and Online Arabic Scripts

| Arabic Script Handwriting | Recognition Rate | Precision |
|---|---|---|
| Offline Category | 68.4% | 62.8% |
| Online Category | 74.6% | 66.3% |

Table 2: The Precision of Combination Online and Offline Arabic Scripts

| Arabic Script Handwriting | Precision |
|---|---|
| Highest Lonely System (Online Category) | 66.3% |
| Combination of (Offline and Online Category) | 68.2% |

## VIII. CONCLUSION

The domain of handwriting in the Arabic script present unprecedented technical challenges and has been addressed more recently than other domain. The Arabic script handwriting recognition is a very challenging task due to its cursive nature. The recognition of cursive handwriting is still an open problem due to the existence of many difficulties such as the variability of the handwritten styles and shapes, writing skew or slant and the size of the lexicon. In this paper describes a compact online and offline combined handwriting recognizer for Arabic scripts.

In this paper we introduced an Arabic script multiple classifier system for recognizing notes written on a Starboard. The Arabic script recognizers are all based on Hidden Markov Models but vary in the way of preprocessing and normalization. Using online and offline combine the output sequences of the recognizers, we enhancement aligned the Arabic word series using a norm string matching algorithm. After that the each output position the word with the most occurrences has been used as the final outcome. We could statistically valued increase the precision by 1.9%. Therefore the proposed technique is also of perforce step towards Arabic script recognition, person identification, personality determination where input data is processed from all perspectives.

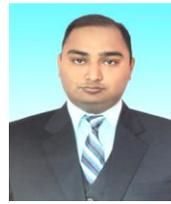

**Dr. Firoj Parwej** Assistant Professor in the Department of Computer Science, Jazan University, Jazan , Kingdom of Saudi Arabia (KSA). He has authored a number of different journal and paper. His research interests include Soft Computing, Artificial Neural Network, Machine Learning, Pattern Matching & Recognition, Artificial Intelligence, Image Processing, Fuzzy Logic, Network and Database. He is a member of IEEE.